\documentclass[11pt,a4paper]{article}
\usepackage[hyperref]{ranlp2023}
\usepackage{times}
\usepackage{latexsym}
\usepackage{lipsum}
\usepackage{graphicx}
\usepackage{scalefnt}
\usepackage{url}
\usepackage{multicol}
\usepackage{subfig}
\usepackage{latexsym}
\usepackage{microtype}
\usepackage{indentfirst}
\usepackage{multirow}
\usepackage{hyperref}
\aclfinalcopy

\usepackage{microtype}

\title{Predicting Sentence-Level Factuality of News and Bias of Media Outlets}

\author{Francielle Vargas{$^1$$^,$$^2$}, Kokil Jaidka{$^3$}, Thiago A. S. Pardo{$^1$}, Fabrício Benevenuto{$^2$} \\
{$^1$}Institute of Mathematical and Computer Sciences, University of São Paulo, Brazil \\
{$^2$}Computer Science Department, Federal University of Minas Gerais, Brazil\\
{$^3$}Centre for Trusted Internet and Community, National University of Singapore, Singapore\\
 \texttt{francielleavargas@usp.br, jaidka@nus.edu.sg} \\\texttt{taspardo@icmc.usp.br, fabricio@dcc.ufmg.br}}
 
\date{}

\begin{document}
\maketitle
\begin{abstract}

Automated news credibility and fact-checking at scale require accurate prediction of news factuality and media bias. This paper introduces a large sentence-level dataset, titled FactNews\footnote{\small\url{https://github.com/franciellevargas/FactNews}}, composed of 6,191 sentences expertly annotated according to factuality and media bias definitions proposed by AllSides. We use the FactNews to assess the overall reliability of news sources by formulating two text classification problems for predicting sentence-level factuality of news reporting and bias of media outlets. Our experiments demonstrate that biased sentences present a higher number of words compared to factual sentences, besides having a predominance of emotions. Hence, the fine-grained analysis of subjectivity and impartiality of news articles showed promising results for predicting the reliability of the media outlets. Finally, due to the severity of fake news and political polarization in Brazil, and the lack of research for Portuguese, both dataset and baseline were proposed for this language\footnote{\small Proceedings of the 14th International Conference on Recent Advances in Natural Language Processing. \url{https://aclanthology.org/2023.ranlp-1.127}.}.
\end{abstract}

\section{Introduction}
\noindent Automated fact-checking and news credibility have become undoubtedly an important research issue mainly due to the potential for misinformation to spread in the modern media ecosystem \cite{10.1162/tacl_a_00454}. Furthermore, although fake news is spreading on social media, it is necessary a source media where they would have been posted originally. Since websites have published low-credible news in the past, it is likely to happen again \citep{baly-etal-2018-predicting}. Nevertheless, automated news credibility to assist human efforts and increase the understanding of the news ecosystem as a whole still requires urgent improvements \cite{Horneetall2018}.

\begin{table}[!htb]
\centering
\scalefont{0.83}
\begin{tabular}{p{2mm}p{55mm}p{7.6mm}}
\hline
\textbf{N.} & \textbf{Sentence-level news article}  & \textbf{Label}\\
\hline
Title & President \textbf{lowers} Brazil's image with repeated misinformation and does not receive attention from global leaders. & Biased \\
\hline
S1 & President Jair Bolsonaro \textbf{touch a sore point of} Europeans when he pointed out that the increased use of fossil fuels is a \textbf{serious} environmental setback, in his opening speech at the UN General Assembly, Tuesday (20). & Biased \\
\hline
S2 & Germany received criticism from the UN for the investment agreement with Senegal for the production of gas in the African country. & 
Factual\\
\hline
S3 & ``This constitutes a serious setback for the environment'', he said, referring to the Europeans & Quotes\\
\hline
S4 & However, Bolsonaro signed measures contrary to environmental protection during the four years of the Brazilian government. & Factual\\
\hline
S5 & There is \textbf{a huge difference} between speaking at the UN and being heard at the UN. & Biased\\
\hline
\end{tabular}
\caption{Sentence-level factuality and bias prediction.}
\label{tab:annotatednews}
\end{table}

 Nowadays, fact-checking organizations have provided lists of unreliable news articles and media sources \citep{baly-etal-2018-predicting}. Notwithstanding, these are inefficient once they need to be updated faster, besides being a very time-consuming task and requiring domain expertise. 

 A strategy to measure the credibility of news sources had already been done using the distribution of biased news in media outlets. While journalism is tied to a set of ethical standards and values, including truth and fairness, it often strays from impartial facts \cite{allsides2022}. As a result, biased news are produced, which may be correlated with the increasing polarization of media \citep{hamborg-2020-media,Prior2013,Gentzkowetall2010}. Moreover, media outlets play an important role in democratic societies \citep{baly-etal-2020-detect} against sophisticated strategies of misinformation.

The state-of-the-art media bias detection has centered around predicting political-ideological bias (left, center, right) of news media. Most of the proposals use \textit{lexical bias} that is linked to lexical and grammatical cues and typically does not depend on context outside of the sentence. Also, it can be alleviated while maintaining its semantics: polarized words can be removed or replaced, and clauses written in active voice can be rewritten in passive voice \cite{fan-etal-2019-plain}. In the same settings, the definition of \textit{frame bias} \citep{recasens-etal-2013-linguistic} is also used to identify media bias, which occurs when subjective or opinion-based words are applied. In a study proposed by  \citeauthor{fan-etal-2019-plain} (\citeyear{fan-etal-2019-plain}), a frame-based analysis was performed for sentence-level media bias detection. The authors suggest that \textit{informational bias} can be considered a specific form of framing in which there is an intention of influencing the reader’s opinion of an entity \citep{fan-etal-2019-plain}. In this paper, we identify sentence-level media bias according to a guideline proposed by AllSides \cite{allsides2022}, which describes 16 different types of media bias.  

Most researchers address media bias and factuality either at the level of media outlet \citep{baly-etal-2018-predicting} or at the level of individual article \citep{roy-goldwasser-2020-weakly,baly-etal-2020-detect}. Nevertheless, each article comprises multiple sentences, which vary in their embedded bias \cite{lim-etal-2020-annotating}, as well as factuality and quotes, as shown in Table \ref{tab:annotatednews}. Observe that factual sentences are a type of information presented with impartiality, focused on objective facts (e.g. S2, S4). In contrast, biased sentences stray from impartial facts and present the point of view of the journalist (e.g. Title, S1, S5), which may influence readers' perceptions. There are also direct quotes, which are neither biased sentences nor factual sentences (e.g. S3). Therefore, the news media sources may affect the power of swaying public opinion through the practical limitation to impartiality or using deliberate attempts to go against or in favor of something or someone.

Taking advantage of the fact that textual analysis of news articles published by a media outlet is critical for assessing the factuality of its reporting, and its potential bias \cite{baly-etal-2018-predicting}, we tackle both biased and factual sentence prediction by using a strategy that has proved to be effective. In accordance with the literature, we created a new dataset titled \textit{FactNews} composed of 6,191 sentences from 100 news stories totaling 300 documents. The same news story was extracted from three different media outlets. Furthermore, each sentence of the dataset was annotated with three different classes according to factuality and media bias definitions proposed by AllSides \cite{allsides2022}: \textbf{(i) factual spans}, which consists of a type of information presented with impartiality focused on the objective fact or, in other words, they are sentences that describe a fact and are committed to objectivity; \textbf{(ii) biased spans}, specifically biased spans were classified according to 12 types of media bias proposed by AllSides \citep{allsides2022}, which we describe in detail in Section \ref{sec:annotationschema}; additionally, \textbf{(iii) quotes} consist of direct statements often followed by quotation marks that journalists in general use to report the speech of someone involved in the reported event. In this paper, we argue that quotes should be defined differently than biased and factual spans. Furthermore, we trained two different models using fine-tuned BERT. The first model predicts whether the sentence of a given news article is factual or not. The second model predicts whether the sentence of a news article from a given news media outlet is biased or not. As a result, baseline models for sentence-level factuality and sentence-level media bias prediction by BERT fine-tuning were presented in order to provide a more accurate score of the reliability of the entire media source.  
Our contributions may be summarized as follows:
\begin{itemize}
    \item  We focus on an under-explored and surely relevant problem: predicting the factuality of news reporting and bias of media outlets.
    \item We create the first large-scale and manually annotated dataset at the sentence-level for both tasks in Portuguese. The dataset, agreements/disagreements, and code are available, which may facilitate future research.
    \item We present a new annotation schema to identify media bias and factuality, as well as a baseline for the factual sentence prediction task.
    \item We provide data analysis on factual and biased sentences demonstrating the reliability of the proposed annotation schema and models.
\end{itemize}

In what follows, in Section \ref{sec:relatedwork}, related work is presented. Section \ref{sec:factnewsdataset} describes the proposed FactNews dataset, and Section \ref{sec:experiments} our experimental settings. In Section \ref{sec:results}, baseline results for sentence-level factuality and media bias prediction are shown. In Section \ref{sec:conclusion}, conclusions are presented.

\section{Related Work}
\label{sec:relatedwork}

\subsection{News Credibility}
\noindent While the assessing of news has been made mainly by journalists, information analysts, and news consumers, this task has become complex due to the ever-growing number of news sources and the mixed tactics of maliciously false sources and misinformation strategies \citep{Horneetall2018}. News credibility state-of-the-art has been mostly focused on measuring the reliability of news reporting \cite{perez-rosas-etal-2018-automatic,10.1007/978-3-319-44748-3_17} or the entire media outlets \cite{baly-etal-2018-predicting,Horneetall2018,baly-etal-2019-multi}, as well as social media platforms \cite{10.1145/1963405.1963500,10.1145/2806416.2806537} in order to mitigate fake news harmful spreading. Furthermore, as stated by \citeauthor{baly-etal-2018-predicting} (\citeyear{baly-etal-2018-predicting}), estimating the reliability of a news source is relevant not only when fact-checking a claim \citep{10.1145/2983323.2983661,Nguyen_Kharosekar_Lease_Wallace_2018}, nevertheless, it provides a surely contribution in order to tackle article-level tasks such as ``fake news'' detection \cite{de-sarkar-etal-2018-attending,yuan-etal-2020-early,reis2019explainable,10.1007/978-3-030-00671-6_39,vargas-etal-2022-rhetorical,10.14778/2777598.2777603}. 
News credibility information has been studied at different levels \citep{baly-etal-2018-predicting}: (i) claim-level (e.g., fact-checking),
(ii) article-level (e.g., “fake news” detection),
(iii) user-level (e.g., hunting for trolls), and
(iv) medium-level (e.g., source reliability estimation). In this paper, we focus on predicting the factuality of reporting and bias of media outlets at medium-level towards source reliability estimation.
\subsection{Fact-Checking}
\noindent 
According to \citeauthor{10.1162/tacl_a_00454} (\citeyear{10.1162/tacl_a_00454}), fake news detection and fact-checking are different tasks once that fact-checkers focus on assessing news articles and include labeling items based on aspects
not related to veracity, besides other factors—such as the audience reached by the claim, and the intentions and forms of the claim—are often considered, as well as the context of propaganda detection \cite{ijcai2020p672}. Fact-checking state-of-the-art at the claim-level, as claimed by \cite{baly-etal-2018-predicting} mostly uses information extracted from social media, i.e., based on how users comment on the target claim \cite{ribeiroetall2022,baly-etal-2019-multi}, so as to the use of the Web data as information source \cite{10.5555/3504035.3504686,icwsm20jreis,10.5555/3504035.3504686}.

\subsection{Media Bias Detection}

\textbf{Article-level media bias} consists of predicting whether a news article is biased. This task was studied in \cite{Sapiro-Gheiler_2019}. They predicted political ideology using recursive neural networks \cite{iyyer-etal-2014-political}. \citeauthor{baly-etal-2019-multi} (\citeyear{baly-etal-2019-multi}) proposed a multi-task regression framework aiming to predict the trustworthiness and ideology of news media. \citeauthor{liu-etal-2022-politics} (\citeyear{liu-etal-2022-politics}) applied the pre-trained language model for the political domain to characterize political stance. \citeauthor{baly-etal-2020-detect} (\citeyear{baly-etal-2020-detect}) created a model from media sources, such as a shortcut, for predicting ideology using adversarial networks.

\indent
\textbf{Sentence-level media bias} consists of a task aiming to predict whether each sentence of a news report is biased or not.  \citeauthor{fan-etal-2019-plain} (\citeyear{fan-etal-2019-plain}) provided the first sentence-level annotated dataset titled \textit{BASIL}, composed of 300 news articles annotated with 1,727 biased spans and 6,257 non-biased sentences, as well as fine-tuning BERT baseline experiments reaching an F1-Score of 47,27\%. \citeauthor{lim-etal-2020-annotating} (\citeyear{lim-etal-2020-annotating}) created a new dataset titled \textit{biased-sents}, which is composed of 966 sentences from 46 English-language news articles covering four different events. \citeauthor{10.1145/3340531.3412876} (\citeyear{10.1145/3340531.3412876}) proposed a dataset of 2,057 sentences annotated with four labels: hidden assumptions, subjectivity, framing, and bias. \citeauthor{spinde-etal-2021-neural-media} (\citeyear{spinde-etal-2021-neural-media}) provided an annotation-expert project through a new dataset titled \textit{BABE}. This dataset consists of 3,700 sentences balanced among topics and outlets, and a fine-tuned BERT baseline reaching an F1-Score of 80,04\%. Lastly, \citeauthor{Leietall2022}(\citeyear{Leietall2022}) showed that embedded discourse structure for sentence-level media bias effectively increases the recall by 8.27\% - 8.62\%, and precision by 2.82\% - 3.48\%.

\subsection{Factuality of News Reporting} 
\noindent Predicting the factuality of news reporting is definitely an under-explored research topic. This task consists of predicting whether a news report on news media is factual or not. \citeauthor{baly-etal-2018-predicting} (\citeyear{baly-etal-2018-predicting}) studied article-level factuality of news reporting. They proposed a baseline by analyzing textual content (syntactic and semantic) of news reporting given a news media source with features based on sentiment, morality, part-of-speech, etc. The best model obtained 58.02\% at F1-Score. \citeauthor{bozhanova-etal-2021-predicting} (\citeyear{bozhanova-etal-2021-predicting}) studied the factuality of reporting of news media outlets by studying the user attention cycles in their YouTube channels. 

\section{FactNews Dataset}
\label{sec:factnewsdataset}
\noindent We collected, annotated, and released a new dataset titled \textit{FactNews}, which consists of a sentence-level annotated dataset in Brazilian Portuguese that contains 6,191 annotated sentences, as follows: 4,302 sentences annotated as factual spans; 1,389 sentences annotated as quotes, and 558 sentences annotated as biased spans. The entire dataset-building process lasted an average of six months. A dataset overview is shown in Table \ref{tab:datasetanalysis1}. We first selected three different well-known and relevant media outlets in Brazil, and extracted the same news story from each one of them, as shown in Table \ref{tab:examples_news}. 

\begin{table}[!htb]
\centering
\scalefont{0.85}
\begin{tabular}{p{12mm}p{56mm}}
\hline
\textbf{Media} & \textbf{News Reporting} \\
\hline
Folha  & O presidente Jair Bolsonaro colocou o dedo na ferida dos europeus ao apontar que o aumento do uso de combustíveis fósseis é um grave retrocesso ambiental, em seu discurso de abertura da Assembleia-Geral da ONU na manhã desta terça-feira (20). 
\textit{\color{blue}{
President Jair Bolsonaro touch a sore point of Europeans when he pointed out that the increased use of fossil fuels is a serious environmental setback, in his opening speech at the UN General Assembly this Tuesday morning (20) (...)}} 
\\\hline
Estadão & O presidente Jair Bolsonaro encerrou seu discurso na Assembleia-Geral da ONU, nesta terça-feira, 20, afirmando que o povo brasileiro acredita em “Deus, Pátria, família e liberdade”, que tem inspiração no fascismo de Benito Mussolini (1883-1945). 
\textit{\color{blue}{
President Jair Bolsonaro ended his speech at the UN General Assembly, this Tuesday, 20, stating that the Brazilian people believe in “God, Fatherland, family and freedom”, which has by the fascism of Benito Mussolini (1883-1945) (...)}} 
\\\hline
O Globo & O presidente Jair Bolsonaro seguiu o roteiro de campanha em seu discurso na Assembleia Geral da Organização das Nações Unidas (ONU), em Nova York (EUA), e aproveitou para atacar o ex-presidente Luiz Inácio Lula da Silva nesta terça-feira (20)(...) 
\textit{\color{blue}{
President Jair Bolsonaro followed the campaign script in his speech at the General Assembly of the United Nations (UN) in New York (USA), and took the opportunity to attack former president Luiz Inácio Lula da Silva this Tuesday (20 )(...)}}
\\\hline
\end{tabular}
\caption{The same news story was collected from three different Brazilian media outlets, which reports the Jair Bolsonaro (former President) speech at the UN in 2022.}
\label{tab:examples_news}
\end{table}

\subsection{Data Collection}
\noindent As shown in Table \ref{tab:datasetanalysis1}, the proposed \textit{FactNews} was collected from 100 news articles in triples -  the same news story from three different Brazilian media news outlets: Folha de São Paulo\footnote{\url{https://www.folha.uol.com.br/}}, O Globo\footnote{\url{https://oglobo.globo.com/}}, and Estadão\footnote{\url{https://www.estadao.com.br/}}, resulting in 300 documents. 

Furthermore, we used a statistical approach and a search algorithm, in order to collect news related to six different domains (e.g. politics, world, daily, sports, science, and culture) from periods 2006-2007 and 2021-2022. Therefore, in accordance with relevant literature of the area, we selected three news articles from different news outlets related to the same topic or story \cite{spinde-etal-2021-neural-media,baly-etal-2020-detect,fan-etal-2019-plain}. 

\subsection{Data Annotation}

\subsubsection{Annotators Profile}
\noindent In order to ensure the reliability of data annotation, two different annotators, a linguist and a computer scientist from different regions (southeast and northeast) performed the task, both with at least a Ph.D. degree or Ph.D. candidate status. Furthermore, the annotation task was led by an NLP researcher, and the annotators were supported by our annotation schema (see Figure \ref{fig:annotationschema}), and a guideline with rich examples proposed by AllSides.

\subsubsection{Annotation Schema}
\label{sec:annotationschema}
\begin{figure*}[!htbp]
    \centering
    \includegraphics[width=0.85\textwidth]{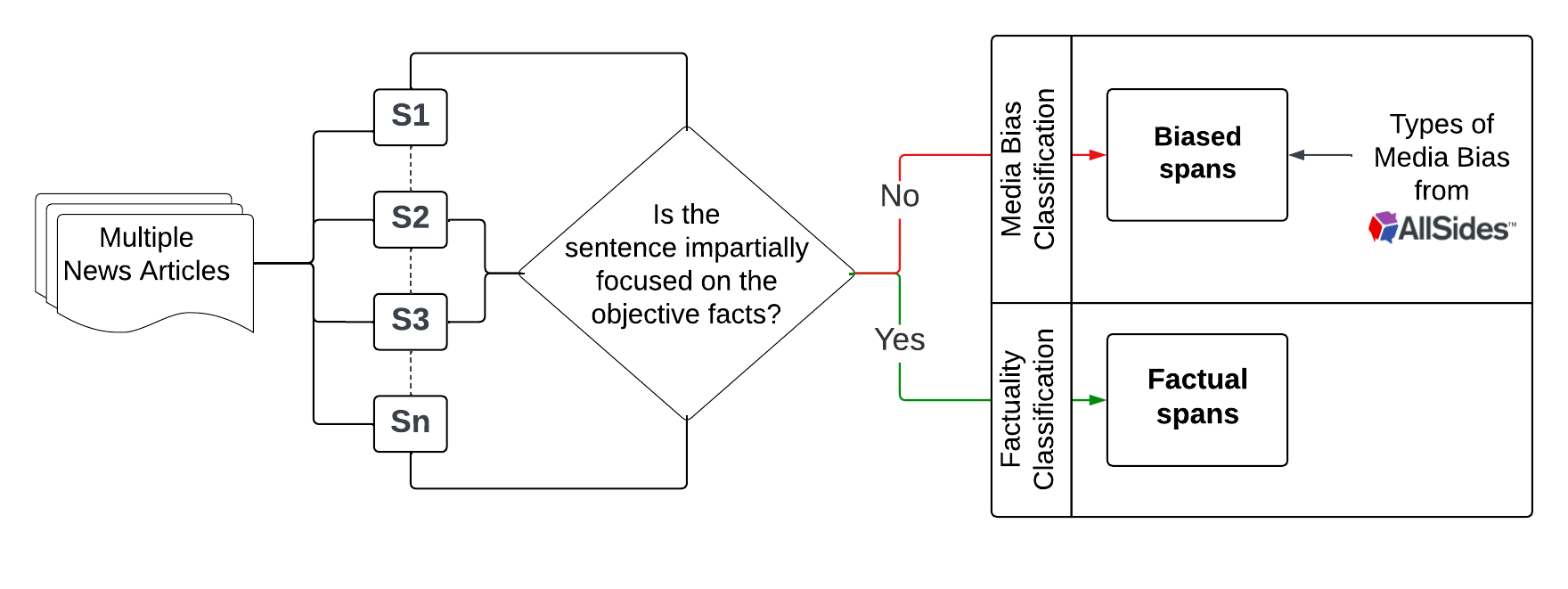}
    \caption{FactNews annotation schema.}
    \label{fig:annotationschema}
\end{figure*}

\noindent Corroborating our objective of classifying factuality and bias at the sentence level, we segmented each one of the 300 news articles in sentences and annotated them according to three different classes: (i) factual spans, (ii) biased spans, and (iii) quotes, as shown in Figure \ref{fig:annotationschema}.

We proposed an expert annotation schema for sentence-level factuality and media bias classification. We first evaluated whether the sentence was committed with impartiality. In other words, whether it presented a type of information focused on objective facts. Whether ``yes'', it should be classified as factual span. Otherwise, it should be classified as a biased span taking into account 12 types of media bias defined by AllSides \citep{allsides2022}, described as follows. We did not consider 4 types (slant, bias by omission, bias by story choice, and photo bias), from the AllSides guidelines\footnote{\url{https://tinyurl.com/3aphktzf}}, once they did not match our sentence-level proposal.  


\begin{enumerate}
    \item \textbf{Spin}: This type of bias consists of vague, dramatic, or sensational language. For example ``President Donald Trump \underline{gloated} over mass layoffs at multiple news outlets on Saturday''. Note that ``gloated'' is evidence of subjective interpretation from the journalist meaning that Trump’s tweet shows he is smug or taking pleasure in the layoffs. 
    
    \item \textbf{Unsubstantiated Claims}: This bias occurs when journalists provide claims in their reporting without including any evidence. For example, ``Sen. Kamala Harris condemned \underline{the violent attack} on actor Jussie Smollett, calling it an attempted modern-day lynching''. 
    
    \item \textbf{Opinion Statements Presented as Facts}: In this bias, journalists use subjective language or statements under the guise of reporting objectively, which is based on personal opinions, assumptions, beliefs, tastes, preferences, or interpretations. For example, ``The EPA is lifting greenhouse gas limits on coal power plants: \underline{The latest proposal won't stop} the steady decline of the coal industry''. Note that the underlined statement shows the point of view of the journalist.

    \item \textbf{Sensationalism/Emotionalism}: Here, the information is presented in a way that provides a shock or triggers a deep impression. For example, ``If seats that look like this one in Rio de Janeiro are toss-ups in November, it's going to be a \underline{bloodbath}''. 
   
    \item \textbf{Mudslinging/Ad Hominem}: This type of media bias occurs when unfair or insulting things are said about someone in order to damage their reputation. For example, ``Bret Stephens is not a \underline{bedbug}. He is a \underline{delicate snowflake}''. 

    \item \textbf{Mind Reading}: This bias occurs when journalists assume they know what another person thinks, or thinks that the way they see the world reflects the way the world really is. For example, ``Bolsonaro's \underline{hatred of looking foolish} and left party' conviction that \underline{they have a winning hand} is leaving the President with no way out of the stalemate over his gun port legalization.''. 


   \item \textbf{Flowed Logic}: This bias consists of a type of faulty reasoning resulting in misrepresenting people’s opinions or arriving at conclusions that are not justified by the given evidence (e.g. arriving at a conclusion that doesn’t follow from the premise). For example, ``Two-time failed Democratic presidential candidate Hillary Clinton \underline{snubbed} Melania Trump during George H.W. Bush's funeral, \underline{refusing to shake her hand} (...), and an awkward and bitter nod back from Hillary''.
   
   \item \textbf{Omission of Source Attribution}: This bias occurs when a journalist does not back up their claims by linking to the source of that information. For example, when journalists claim ``critics say'' without specific attribution.
   
    \item \textbf{Subjective Qualifying Adjectives}: Journalists can reveal this bias when they include subjective, qualifying adjectives in front of specific words or phrases. For example, ``Rep. Madison Cawthorn issues \underline{sinister warning} to anyone opposing Him. The \underline{extremist republican} ranted about liberals trying to make people ``sexless''''.   Note that subjective qualifiers are closely related to \textit{spin words} and phrases once they obscure the objective truth and insert subjectivity.
    
    \item \textbf{Word Choice}: This bias occurs when words and phrases are loaded with political implications. Therefore, the words or phrases a media outlet uses can reveal its perspective or ideology. Examples of Polarizing Word Choices: ``pro-choice | anti-choice'', ``gun rights | gun control'', ``riot | protest'', ``illegal immigrants | migrants''.

    \item \textbf{Negativity Bias}: Journalists can emphasize bad or negative news, or frame events in a negative light. For example, news articles related to death, violence, turmoil, and struggle, tend to obtain more attention and elicit more shock, and fear. As a result, we keep reading the news, in order to know more on this issue.

    \item \textbf{Elite v. Populist Bias}: Journalists can defer to the beliefs, viewpoints, and perspectives of people who are part of society's most prestigious or not prestigious. Furthermore, Elite/populist bias has a geographic component. For example, ``The FDA turned a blind eye or colluded with unbelievable harms revealed in the Pfizer documents, so the FDA can't be trusted. \underline{The CDC can't be trusted}''. Here, the journalist pushes back against the elite government, saying they can't be trusted.
    
\end{enumerate}


\subsubsection{Annotation Evaluation}
\noindent We computed the inter-annotator agreement score using Cohen's kappa  \cite{sim2005kappa}. We obtained a kappa score of 82\%. We also analyzed the matrix of agreements and disagreements among annotators for each class (e.g. factual, biased, and quotes). Results are shown in Table \ref{tab:interanalysis}. 

\begin{table}[!htb]
\scalefont{0.70}
\centering
\begin{tabular}{ll|cccc}
\hline
\multicolumn{2}{l|}{\multirow{2}{*}{\textbf{FactNews Dataset}}}               & \multicolumn{3}{c|}{\textbf{Annotator 1}}                                                                           & \multicolumn{1}{l}{\multirow{2}{*}{\textbf{Total}}} \\ \cline{3-5}
\multicolumn{2}{l|}{}                                      & \multicolumn{1}{l|}{\textit{Factual}} & \multicolumn{1}{l|}{\textit{Biased}} & \multicolumn{1}{l|}{\textit{Quotes}} & \multicolumn{1}{l}{}                                \\ \hline
\multicolumn{1}{l|}{\multirow{3}{*}{\textbf{Annotator 2}}} & \textit{Factual} & \multicolumn{1}{c|}{4,211}            & \multicolumn{1}{c|}{27}              & \multicolumn{1}{c|}{7}               & 4,245                                               \\ \cline{2-6} 
\multicolumn{1}{l|}{}                                      & \textit{Biased}  & \multicolumn{1}{c|}{284}              & \multicolumn{1}{c|}{261}             & \multicolumn{1}{c|}{1}               & 546                                                 \\ \cline{2-6} 
\multicolumn{1}{l|}{}                                      & \textit{Quotes}  & \multicolumn{1}{c|}{138}              & \multicolumn{1}{c|}{6}               & \multicolumn{1}{c|}{1,256}           & 1,400                                               \\ \hline
\multicolumn{2}{l|}{\textbf{Total}}                                           & \multicolumn{1}{c|}{4,633}            & \multicolumn{1}{c|}{294}             & \multicolumn{1}{c|}{1,264}           & \textbf{6,191}                                      \\ \hline
\multicolumn{2}{l|}{\textbf{Kappa}}                                           & \multicolumn{4}{c}{\textbf{0.82}}                                                                                                                                         \\ \hline
\end{tabular}\caption{Inter-annotator agreement by Kappa.}
\label{tab:interanalysis}
\end{table}

Observe that two annotators, a linguist (expert in media bias) and a computer scientist (non-expert) labeled the FactNews dataset. Moreover, disagreement cases\footnote{\url{https://zenodo.org/record/7868597/}} were also judged by two judges, and three meetings were carried out, in which annotators could discuss and re-evaluate the given labels. 



Furthermore, as shown in Table \ref{tab:interanalysis}, the high values obtained by diagonal lines (e.g. 4,211, 261, 1,256) are indicative of high-human agreement. We also observed that annotator 2, which is a specialist, provides better media bias classification compared with annotator 1, which is a non-specialist. For example, while both annotators agreed on the bias labels with 261 matches between them, 284 labels were classified by annotator 1 as ``factual'' and by annotator 2 (specialist) as ``biased''. These cases were mostly decided by judges as being ``biased''.

\subsection{Data Analysis}
\noindent Table \ref{tab:datasetanalysis1} shows the dataset statistics. The FactNews is composed of 6,191 sentences annotated according to three classes: factual spans (4,242), quotes (1,391), and biased spans (558). Most of the sentences (68.51\%) are factual spans, in contrast to quotes (22.52\%) and biased (8.81\%) categories, respectively. Each news article consists of an average of 24.27 sentences of which 14.14 are classified as factual sentences, 7.06 as quotes, and 3.27 as biased sentences. Furthermore, factual sentences contain an average of 20.36 words, biased sentences 22.14 words, and quotes 17.38 words.

Furthermore, biased spans present more words than factual spans in all grammar categories (e.g. nouns, verbs, adjectives), as well as predominance in terms of emotion lexicon. Lastly, the titles of news articles hold 8.36\% bias, 5.33\% quotes, and 86\% of factual sentences. On the other hand, the body of news articles holds 13.35\% bias, 20.38\% quotes, and 66.27\% of factual sentences. 

\begin{figure}[!htbp]
    \centering
    \includegraphics[width=0.23\textwidth]{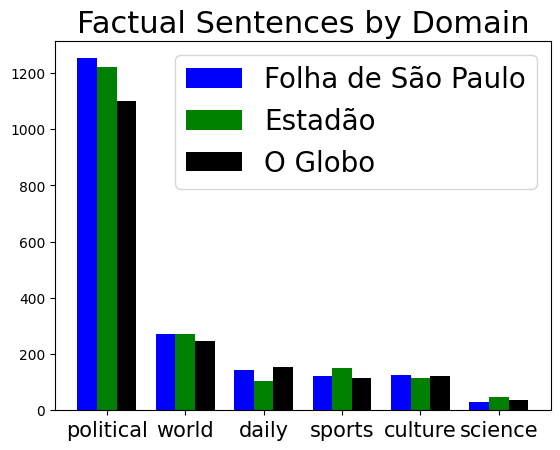}
    \includegraphics[width=0.23\textwidth]{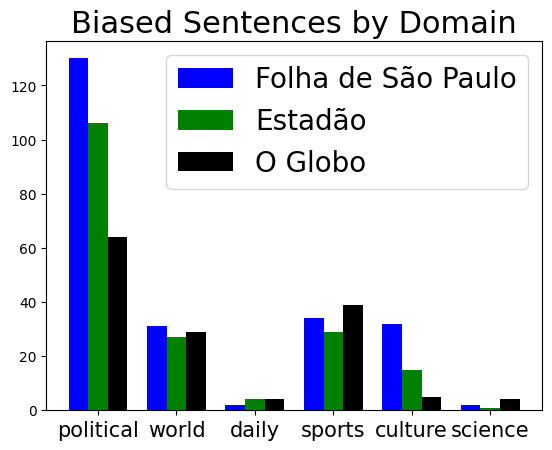}
    \caption{The cross-domain distribution of factual and biased sentences from different media outlets.}
    \label{fig:factual-biased}
\end{figure}

In Figure \ref{fig:factual-biased}, we also show the distribution of factual and biased sentences across domains according to each media news outlet. Notably, the distribution of factuality is equivalent across different domains. Differently, the distribution of bias varies in accordance with the domain and media outlet. Considering the labels across domains, 62.55\% are related to politics; 14.21\% world; 7.14 sport; 6.67 daily; 6.65 culture; and 1.98\% science. 

\begin{table*}[!htb]
\centering
\scalefont{0.70}
\begin{tabular}{|cc|ccc|ccc|ccc|c|}
\hline
\multicolumn{2}{|c|}{\multirow{2}{*}{Description}}                                                                                     & \multicolumn{3}{c|}{\textbf{Folha de São Paulo}}     & \multicolumn{3}{c|}{\textbf{Estadão}}                & \multicolumn{3}{c|}{\textbf{O Globo}}                & \multirow{2}{*}{\textbf{All}} \\ \cline{3-11}
\multicolumn{2}{|c|}{}                                                                                          & \textbf{factual} & \textbf{quotes} & \textbf{biased} & \textbf{factual} & \textbf{quotes} & \textbf{biased} & \textbf{factual} & \textbf{quotes} & \textbf{biased} &                               \\ \hline
\multicolumn{2}{|c|}{\#Articles}                                                                                                       & \multicolumn{3}{c|}{100}                             & \multicolumn{3}{c|}{100}                             & \multicolumn{3}{c|}{100}                             & 300                           \\ \cline{3-12} 
\multicolumn{2}{|c|}{\#Sentences}                                                                                                      & 1,494            & 450             & 231             & 1,428            & 483             & 182             & 1,320            & 458             & 145             & 6191                          \\
\multicolumn{2}{|c|}{\#Words}                                                                                                          & 30,374           & 7,946           & 5,177           & 30,589           & 8,504           & 4,002           & 25,505           & 7,740           & 3,195           & 123,032                       \\
\multicolumn{2}{|c|}{Avg Sentences/Article}                                                                                             & 14.94            & 7.03            & 3.78            & 14.28            & 7.00            & 3.19            & 13.20            & 7.15            & 2.84            & 8.15                          \\
\multicolumn{2}{|c|}{Avg Words/Sentences}                                                                                              & 20.33            & 17.65           & \textbf{22,41}  & 21,45            & 17,60           & \textbf{21,98}  & 19,32            & 16,89           & \textbf{22,03}  & 19,96                         \\ \hline
\multicolumn{1}{|c|}{\multirow{2}{*}{\textbf{Body/Title}}}                                                      & \textit{Body}        & 1,337            & 440             & 207             & 1,218            & 473             & 162             & 1,089            & 441             & 131             & 5,498                         \\
\multicolumn{1}{|c|}{}                                                                                          & \textit{Title}       & 157              & 10              & 24              & 210              & 10              & 20              & 231              & 17              & 14              & 693                           \\ \hline
\multicolumn{1}{|c|}{\multirow{6}{*}{\textbf{Domains}}}                                                         & \textit{Political}   & 912              & 340             & 130             & 870              & 352             & 106             & 748              & 351             & 64              & 3,873                         \\
\multicolumn{1}{|c|}{}                                                                                          & \textit{World}       & 224              & 48              & 31              & 224              & 49              & 27              & 216              & 32              & 29              & 880                           \\
\multicolumn{1}{|c|}{}                                                                                          & \textit{Sports}      & 100              & 23              & 34              & 124              & 25              & 29              & 98               & 18              & 39              & 490                           \\
\multicolumn{1}{|c|}{}                                                                                          & \textit{Daily}  & 132              & 11              & 2               & 98               & 7               & 4               & 148              & 7               & 4               & 413                           \\
\multicolumn{1}{|c|}{}                                                                                          & \textit{Culture}     & 98               & 26              & 32              & 72               & 42              & 15              & 77               & 45              & 5               & 412                           \\
\multicolumn{1}{|c|}{}                                                                                          & \textit{Science}     & 28               & 2               & 2               & 40               & 8               & 1               & 33               & 5               & 4               & 123                           \\ \hline
\multicolumn{1}{|c|}{\multirow{6}{*}{\textbf{\begin{tabular}[c]{@{}c@{}}Part-of-speech \\ (Avg)\end{tabular}}}} & \textit{Noun}        & 4.85             & 4.09            & \textbf{5.72}   & 5.21             & 4.12            & \textbf{5.60}   & 4.59             & 3.82            & \textbf{5.19}   & 4.79                          \\
\multicolumn{1}{|c|}{}                                                                                          & \textit{Verb}        & 2.20             & 2.55            & \textbf{2.60}   & 2.28             & 2.51            & \textbf{2.53}   & 2.00             & 2.44            & \textbf{2.57}   & 4.18                          \\
\multicolumn{1}{|c|}{}                                                                                          & \textit{Adjective}   & 1.03             & 1.03            & \textbf{1.32}   & 1.11             & 1.08            & \textbf{1.32}   & 0.94             & 0.97            & \textbf{1.48}   & 1.14                          \\
\multicolumn{1}{|c|}{}                                                                                          & \textit{Adverb}      & 0.67             & 0.82            & \textbf{0.93}   & 0.67             & 0.94            & \textbf{0.90}   & 0.59             & 0.90            & \textbf{0.94}   & 0.81                          \\
\multicolumn{1}{|c|}{}                                                                                          & \textit{Pronoun}     & 0.52             & 1.02            & \textbf{0.73}   & 0.51             & 0.97            & \textbf{0.56}   & 0.47             & 0.90            & \textbf{0.59}   & 0.69                          \\
\multicolumn{1}{|c|}{}                                                                                          & \textit{Conjunction} & 0.51             & 0.55            & \textbf{0.61}   & 0.54             & 0.57            & \textbf{0.73}   & 0.51             & 0.88            & \textbf{0.70}   & 0.62                          \\ \hline
\multicolumn{1}{|c|}{\multirow{6}{*}{\textbf{\begin{tabular}[c]{@{}c@{}}Emotion\\ (Avg)\end{tabular}}}}         & \textit{Happiness}   & 0.12             & 0.22            & \textbf{0.20}   & 0.16             & 0.28            & \textbf{0.26}   & 0.13             & 0.28            & \textbf{0.22}   & 0.20                          \\
\multicolumn{1}{|c|}{}                                                                                          & \textit{Disgust}     & 0.03             & 0.06            & \textbf{0.05}   & 0.04             & 0.06            & \textbf{0.03}   & 0.04             & 0.04            & \textbf{0.04}   & 0.04                          \\
\multicolumn{1}{|c|}{}                                                                                          & \textit{Fear}        & \textbf{4.18}    & \textbf{3.80}   & \textbf{4.63}   & \textbf{4.41}    & \textbf{3.77}   & \textbf{4.56}   & \textbf{4.05}    & \textbf{3.60}   & \textbf{4.50}   & 4.16                          \\
\multicolumn{1}{|c|}{}                                                                                          & \textit{Anger}       & 0.05             & 0.06            & \textbf{0.13}   & 0.07             & 0.07            & \textbf{0.12}   & 0.06             & 0.08            & \textbf{0.20}   & 0.09                          \\
\multicolumn{1}{|c|}{}                                                                                          & \textit{Surprise}    & 0.01             & 0.03            & \textbf{0.03}   & 0.01             & 0.03            & \textbf{0.05}   & 0.01             & 0.02            & \textbf{0.01}   & 0.02                          \\
\multicolumn{1}{|c|}{}                                                                                          & \textit{Sadness}     & \textbf{5.86}    & \textbf{5.71}   & \textbf{6.52}   & \textbf{6.17}    & \textbf{5.55}   & \textbf{6.48}   & \textbf{5.56}    & \textbf{5.40}   & \textbf{6.19}   & 5.93                          \\ \hline
\multicolumn{1}{|c|}{\multirow{3}{*}{\textbf{\begin{tabular}[c]{@{}c@{}}Polarity\\ (Avg)\end{tabular}}}}        & \textit{Positive}    & 2.41             & 3.25            & 2.93            & 2.55             & 3.22            & 2.95            & 2.26             & 3.26            & 2.96            & 2.86                          \\
\multicolumn{1}{|c|}{}                                                                                          & \textit{Negative}    & 0.05             & 0.06            & 0.05            & 0.07             & 0.10            & 0.09            & 0.06             & 0.07            & 0.06            & 0.06                          \\
\multicolumn{1}{|c|}{}                                                                                          & \textit{Neutral}     & 9.55             & 9.77            & 10.93           & 9.92             & 9.52            & 11.03           & 8.91             & 9.28            & 10.56           & 9.94                          \\ \hline
\end{tabular}
\caption{FactNews dataset statistics.}
\label{tab:datasetanalysis1}
\end{table*}

\section{Baseline Experiments}
\label{sec:experiments}
\subsection{Motivations and Goals}
\noindent As mentioned before, news credibility analysis and fact-checking are both time-consuming tasks. Furthermore, with the amount of new information that appears and the speed with which it spreads, manual validation is insufficient \cite{10.1162/tacl_a_00454}. Nevertheless, automated approaches present several challenges, since automated trustworthiness analysis is a technically complex issue, besides involving a wide variety of ethical dilemmas.

Instead of analyzing the veracity of news articles, in this paper, we are interested in the fine-grained characterization of the entire media outlet by predicting the factuality of news reporting and bias of media outlets for source reliability estimation. 

We aim to predict sentence-level media bias and factuality by analyzing different types of media bias and journalist factuality definitions, both proposed by AllSides \cite{allsides2022}. Specifically, we first built the state-of-art media bias detection models. Secondly, a baseline sentence-level factuality detection model was proposed by analyzing the subjectivity and impartiality of text content. As a result, we hope to explain more accurately the overall reliability of the entire news media source.


\subsection{Model Architecture}
\noindent First of all, we argue that factual spans contain a type of information that deals with facts, hence it is impartially focused on objective facts. In contrast, non-factual information contains a type of information presented subjectively (with partiality) that often strays from objective facts. Taking into account this premise, we describe both model's sentence-level media bias and factuality, as follows:

\textbf{Sentence-Level Media Bias Model}: We implemented the state-of-the-art sentence-level media bias models \cite{fan-etal-2019-plain} on the FactNews dataset. Our model for media bias uses a binary class variable composed of biased spans (558 labels) versus unbiased spans (558 labels).  
 
\textbf{Sentence-Level Factuality Model}: We hypothesize that the factuality of news reporting may be predicted by analyzing the subjectivity and impartiality of text content, which is inspired by \citeauthor{baly-etal-2018-predicting}, (\citeyear{baly-etal-2018-predicting}). Since factual sentences are impartially focused on objective facts, in contrast to the biased ones that are partially presented and focused on subjective interpretations, we built a model to predict sentence-level factuality based on aspects of subjectivity and impartiality. Finally, once both biased spans and quotes present evidence of subjective interpretation of facts \cite{hu2023}, our sentence-level factuality model is composed of a binary class variable from biased spans and quotes (1,949 labels) versus factual spans (1,949 labels).


\subsection{Learning Methods and Features Set}
\noindent In data preparation, we segmented sentences using the spaCy library and only special characters were removed. As learning method, we used the SVM with linear kernel. We split our data into train (90\%), and test (10\%), and applied the 10-fold cross-validation. We also used the undersampling \citep{witten2016data} to balance the classes. Finally, a robust set of experiments was performed using four model architectures inspired by \citeauthor{baly-etal-2018-predicting} (\citeyear{baly-etal-2018-predicting}), which we describe in detail as follows: 

\textbf{BERT fine-tuning}: We used the best BERT fine-tuned model by Keras, held batch size at 64, maximum of 500 features, learning rate at 2e-05 and number of epochs at 4. 

\textbf{Subjective-lexicons}: We evaluated a BoW using features extracted from sentiment and emotion lexicons \cite{WordNetAffectBR2008}, which present semantic polarity and emotion types. 

\textbf{Part-of-speech (POS)}: We evaluated a BoW using features based on POS, more precisely, noun, verb, adjective, adverb, pronoun, and conjunctions, which was supported by the spaCy tagging. 

\textbf{TF-IDF}: Baseline vector space model. 
    
\section{Results and Discussion}
\label{sec:results}
\noindent Table \ref{tab:evaluation} summarizes the performance of the models. We further provide a comparison of results in Table \ref{tab:comparison}. The best model for sentence-level factuality prediction obtained 88\% of F1-Score. For sentence-level media bias prediction, the best model obtained 67\% of F1-score. Notably, the part-of-speech model presented competitive results for both tasks in contrast to the subjective lexicons, which obtained poor results for both tasks. 

\begin{table}[!htb]
\scalefont{0.70}
\centering
\begin{tabular}{lccc}
\hline
\multicolumn{1}{c}{\textbf{Sentence-Level Factuality}} & \textbf{Precision} & \textbf{Recall} & \textbf{F1-Score} \\ \hline
BERT fine-tuning                                       & 0.89               & 0.89            & \textbf{0.88}     \\
Part-of-speech                                         & 0.77               & 0.77            & 0.76              \\
TF-IDF                                                 & 0.81               & 0.69            & 0.66              \\
Polarity-lexicon                                       & 0.63               & 0.62            & 0.62              \\
Emotion-lexicon                                        & 0.61               & 0.61            & 0.61              \\ \hline
\multicolumn{1}{c}{\textbf{Sentence-Level Media Bias}} & \textbf{Precision} & \textbf{Recall} & \textbf{F1-Score} \\ \hline
BERT fine-tuning                                       & 0.70               & 0.68            & \textbf{0.67}     \\
Part-of-speech                                         & 0.67               & 0.66            & 0.66              \\
Polarity-lexicon                                       & 0.50               & 0.50            & 0.50              \\
Emotion-lexicon                                        & 0.53               & 0.52            & 0.50              \\
TF-IDF                                                 & 0.78               & 0.58            & 0.48              \\ \hline
\end{tabular}
\caption{Sentence-level factuality and bias prediction.}
\label{tab:evaluation}
\end{table}

\subsection{Comparing Results}
\noindent While a direct comparison is unfair (as the authors use different datasets), it offers an idea of the general performance, as shown in Table \ref{tab:comparison}. Note that although it only offers an idea of the general performance, our sentence-level factuality prediction model (88\%) significantly outperforms the article-level factuality prediction baseline (58\%).

\begin{table}[!htb]
\scalefont{0.66}
\centering
\begin{tabular}{lcccc}
\hline
\multicolumn{5}{c}{\textbf{Sentence-Level Media Bias Prediction}}                                           \\ \hline
\textbf{Datasets}   & \textbf{Lang} & \textbf{Docum.}                  & \textbf{Sent.} & \textbf{F1-Score} \\ \hline
BASIL (baseline)    & En            & 300 news                         & 7,984          & \textbf{0.47}             \\
Biased-sents        & En            & 46 news                          & 966            & -                 \\
BABE                & En            & 100 news                         & 3,700          & 0.80     \\
FactNews            & Pt            & 300 news                         & 6,191          & \textbf{0.67}     \\ \hline
\multicolumn{5}{c}{\textbf{Sentence-Level Factuality Prediction}}                                           \\ \hline
FactNews (baseline) & Pt            & 300 news                         & 6,191          & \textbf{0.88}     \\ \hline
\multicolumn{5}{c}{\textbf{Article-Level Factuality Prediction}}                                            \\ \hline
MBFC  (baseline)    & En            & \multicolumn{1}{l}{1,066 medias} & -              & \textbf{0.58}              \\
MBFC corpus         & En            & 489 medias                       & -              & 0.76*        \\ \hline
\end{tabular}
\caption{Result analysis in comparison with literature.}
\label{tab:comparison}
\end{table}

\section{Conclusions}
\label{sec:conclusion}
\noindent Since low-credibility media outlets may potentially be targeted for the spreading of misinformation, we study the factuality of news reporting and bias of media outlets at the sentence-level for fine-grained source reliability estimation. We further provide a new data resource and baselines for Brazilian Portuguese low-resourced language. We first created a large and manually-annotated dataset for sentence-level factuality and media bias prediction. Then, we provided a detailed data analysis, demonstrating the reliability of the annotation schema and models. Finally, baseline models for sentence-level factuality and media bias prediction by BERT were presented in order to provide an accurate score of the reliability of the entire news media. Results also showed that biased spans are more numerous in words and emotions compared to factual spans. Moreover, media outlets presented different proportions of bias, and its distribution in news articles may vary according to the domain, in contrast to factual spans. We also concluded that expert annotators are more successful to identify media bias.

\section*{Acknowledgements}
This project was partially funded by the SINCH, FAPESP, FAPEMIG, and CNPq, as well as the Ministry of Science, Technology and Innovation, with resources of Law N. 8.248, of October 23, 1991, within the scope of PPI-SOFTEX, coordinated by Softex and published as Residence in TIC 13, DOU 01245.010222/2022-44.



\bibliographystyle{acl_natbib}
\bibliography{ranlp2023}


\end{document}